\documentclass[11pt,a4paper]{article}
\usepackage{authblk}

\usepackage[hyperref]{emnlp-ijcnlp-2019}
\usepackage{times}
\usepackage{latexsym}
\usepackage{graphicx}

\usepackage{url}

\usepackage{booktabs}
\usepackage{multirow}
\usepackage{makecell}
\usepackage{cleveref}
\crefformat{section}{\S#2#1#3} % see manual of cleveref, section 8.2.1
\crefformat{subsection}{\S#2#1#3}
\crefformat{subsubsection}{\S#2#1#3}

\aclfinalcopy % Uncomment this line for the final submission

\title{A Search-based Neural Model for Biomedical Nested and Overlapping Event Detection}

\author[1,4]{Kurt Espinosa}
\author[2,3]{Makoto Miwa}
\author[1]{Sophia Ananiadou}
\affil[1]{National Centre for Text Mining, School of Computer Science, The University of Manchester, UK}
\affil[2]{Toyota Technological Institute, Nagoya, 468-8511, Japan}
\affil[3]{Artificial Intelligence Research Center (AIRC), \authorcr
\textnormal{\normalsize National Institute of Advanced Industrial Science and Technology (AIST), Japan}}
\affil[4]{Department of Computer Science, University of the Philippines Cebu, Cebu City, Philippines}
\affil[ ]{\tt \{kurtjunshean.espinosa, sophia.ananiadou\}@manchester.ac.uk}
\affil[ ]{\tt makoto-miwa@toyota-ti.ac.jp}

\date{}

\begin{document}
\maketitle
\begin{abstract}
We tackle the nested and overlapping event detection task and propose a novel search-based neural network (SBNN) structured prediction model that treats the task as a search problem on a relation graph of trigger-argument structures. 
Unlike existing structured prediction tasks such as dependency parsing, the task targets to detect DAG structures, which constitute events, from the relation graph.
We define actions to construct events and use all the beams in a beam search to detect all event structures that may be overlapping and nested.
The search process constructs events in a bottom-up manner while modelling the global properties for nested and overlapping structures simultaneously using neural networks. We show that the model achieves performance comparable to the state-of-the-art model Turku Event Extraction System (TEES) on the BioNLP Cancer Genetics (CG) Shared Task 2013 without the use of any syntactic and hand-engineered features. Further analyses on the development set show that our model is more computationally efficient while yielding higher F1-score performance.
\end{abstract}

\begin{figure}[t!]
\includegraphics[scale=0.4,width=\linewidth]{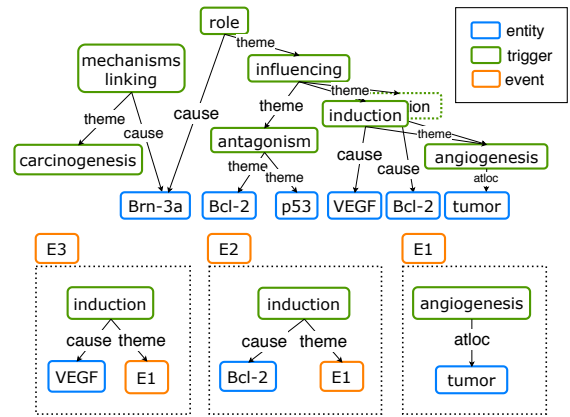}
\caption{Top: A DAG-structured relation graph (topmost) from the sentence ``\textit{Looking for mechanisms linking Brn-3a to carcinogenesis, we discuss the role of this transcription factor in influencing Bcl-2/VEGF induction of tumor angiogenesis, ...}'' from BioNLP'13 CG Shared Task~\cite{pyysalo2015overview}. Bottom: A pair of overlapping and nested events $(E2,E3)$ extracted from the graph with their shared argument event, a flat event $(E1)$.}
\label{example}
\end{figure}

\section{Introduction}
Nested and overlapping event structures, which occur widely in text, are important because they can capture relations between events such as causality, e.g., a ``production'' event is a consequence of a ``discovery'' event, which in turn is a result of an ``exploration'' event. Event extraction involves the identification of a trigger and a set of its arguments in a given text. Figure~\ref{example} shows an example of a nested and overlapping event structure in the biomedical domain. The relation graph (topmost) forms a directed acyclic graph (DAG) structure~\citep{mcclosky2011event} and it encapsulates 15 event structures. It contains nested event structures such as \textit{E2,E3} because one of their arguments, in this case \textit{E1}, is an event. Specifically, \textit{E1} is a flat event since its argument is an entity. Moreover, \textit{E2} and \textit{E3} are also overlapping events (explicitly shown in the relation graph having two \textit{induction} triggers) because they share a common argument, \textit{E1}.

State-of-the-art approaches to event extraction in the biomedical domain are pipeline systems~\cite{bjorne2018biomedical,miwa2013wide} that decompose event extraction into simpler tasks such as: 
i) \textit{trigger/entity detection}, 
which determines which words and phrases in a sentence potentially constitute as participants of an event, 
ii) \textit{relation detection}, 
which finds pairwise relations between triggers and arguments, 
and iii) \textit{event detection}, which combines pairwise relations into complete event structures. 
% In this work, we focus on \textit{event detection} since the current methods do not consider nested and overlapping structures simultaneously during learning. Treating them simultaneously helps the model avoid inferring wrong causality relations between entities. 
Joint approaches have also been explored~\cite{rao2017biomedical,riedel2011robust,vlachos2012biomedical,venugopal-EtAl:2014:EMNLP2014}, but they focus on finding relation graphs and detect events with rules. \citet{mcclosky2011event} treats events as dependency structures by constraining event structures to map to trees, thus their method cannot represent overlapping event structures. Other neural models in event extraction are in the general domain~\cite{feng2016language,nguyen2015event,chen2015event,nguyen2016joint}, but they used the ACE2005 corpus which does not have nested events~\cite{miwa2014comparable}. Furthermore, there are some efforts on applying transition-based methods on DAG structures in dependency parsing, e.g., \cite{C08-1095,AAAI1816549}, however, they do not consider overlapping and nested structures. 

We present a novel search-based neural \textit{event detection} model that detects overlapping and nested events with beam search by formulating it as a structured prediction task for DAG structures. Given a relation graph of trigger-argument relations, our model detects nested events by searching a sequence of actions that construct event structures incrementally in a bottom-up manner. 
We focus on \textit{event detection} since the existing methods do not consider nested and overlapping structures as a whole during learning. Treating them simultaneously helps the model avoid inferring wrong causality relations between entities.
Our model detects overlapping events by maintaining multiple beams and detecting events from all the beams, in contrast to existing transition-based methods~\cite{nivre2003efficient,nivre2006inductive,chen2014fast,dyer2015transition,andor2016globally,vlachos2012biomedical}. We define an LSTM-based neural network model that can represent nested event structure to choose actions. %Finally, we evaluated our model on the BioNLP Cancer Genetics (CG) Shared Task 2013~\cite{pyysalo2013overview} using the predicted relations of the state-of-the-art TEES system~\cite{bjorne2018biomedical}. 

We show that our event detection model achieves performance comparable to the event detection module of the state-of-the-art TEES system~\cite{bjorne2018biomedical} on the BioNLP CG Shared Task 2013~\cite{pyysalo2013overview} without the use of any syntactic and hand-engineered features. Furthermore, analyses on the development set show our model performs fewer number of classifications in less time.

\begin{figure}[t!]
\centering
    \includegraphics[width=\linewidth, keepaspectratio]{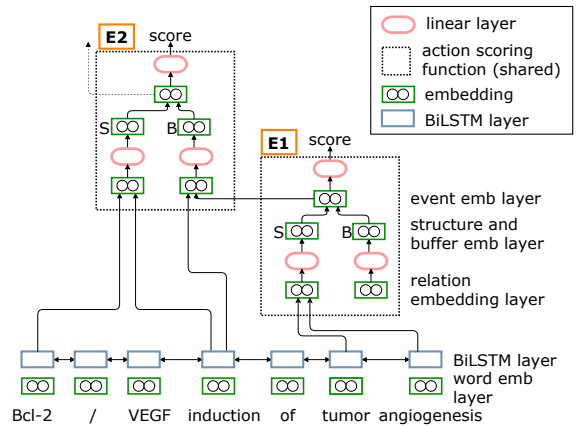}
    \caption{An illustration of the proposed neural model detecting event structures in a bottom-up manner, where E1 event representation becomes an argument to E2 event structure on the example sentence used in Figure~\ref{example}.}
    \label{fig:nn}
\end{figure}

\section{Model}

We describe our search-based neural network (SBNN) model that constitutes events from a relation graph by structured prediction.
SBNN resembles an incremental transition-based parser~\cite{nivre2006inductive}, but the search order, actions and representations are defined for DAG structures. 
We first discuss how we generate the relation graph in \S \ref{relgraph}, then describe the structured prediction algorithm in \S \ref{searchprocess} and the neural network in \S \ref{nnmodel} and lastly, we explain the training procedure in \S \ref{training}. 

\subsection{Relation Graph Generation}
\label{relgraph}
% Since we compare our model with the state-of-the-art TEES event detection module~\cite{bjorne2018biomedical}, we train our model with the same input as theirs. Concretely, we merge the predicted relations from TEES with the pairwise relations decomposed from the gold events.
To train our model, we use the predicted relations merged with pairwise relations decomposed from the gold events. We then generate a relation graph from the merged relations. During inference, the relation graph is generated only from the predicted relations. Figure \ref{example} shows an example of the generated DAG-structured relation graph. 

\subsection{Structured Prediction for DAGs}
\label{searchprocess}
We represent event structures with DAG structures and find them in a relation graph. 
%A DAG structure, unlike a tree structure, can have multiple paths between two nodes. 
Our model performs beam search on relation graphs by choosing actions to construct events. Unlike existing beam search usage where they only choose the best path in the beam, e.g., ~\cite{nivre2006inductive}, we use \textit{all} the beams to predict event structures, which enables the model to predict overlapping and nested events.

Event structures are searched and fixed for each trigger of the relation graph in a bottom-up manner. The model predicts the flat events first and then the representations of the flat events become the argument of the nested events. Figure \ref{fig:nn} shows the proposed neural model (described in detail in \S \ref{nnmodel}), which illustrates that the event representation of event $E1$ becomes the argument to event $E2$. If the flat event is not detected, its nested event will not be detected consequently. When this happens, the search process stops.

% \begin{figure}[t!]
%   \centering
%     \includegraphics[width=\linewidth, keepaspectratio]{graphics/snapshot.pdf}
%     \caption{An illustration of the search procedure with the sequence of actions taken to construct an event structure.}
%     \label{fig:search}
% \end{figure}

\begin{figure}[t!]
  \centering
    \includegraphics[width=\linewidth, keepaspectratio]{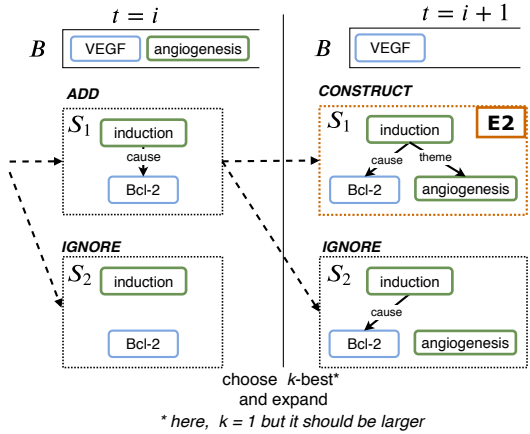}
    \caption{A snapshot of the search procedure within one time step (with $k=1$) on the trigger \textit{induction} which detects event $E2$ after a CONSTRUCT action.}
    \label{fig:search}
\end{figure}

Figure \ref{fig:search} shows an snapshot of the search procedure within one time step as applied to a relation graph with a trigger \textit{induction} to detect event $E2$ (see Figure \ref{example} and \ref{fig:nn}).
%\footnote{No-argument event handling, the initial state and the role types in $B$ are omitted for simplicity.} 
To do the search, the model maintains two data structures: a buffer $B$ which holds a queue of arguments\footnote{A special NONE argument marks the first argument in the buffer to enable detection of no-argument events.} to be processed and a structure $S$ which contains the partially built event structure. The initial state is composed of the buffer $B$ with all the arguments and a structure $S$ empty. At each succeeding time step, the model applies a set of predefined actions to each argument and uses a neural network to score those actions. Processing completes when $B$ is empty and $S$ contains all the arguments for the trigger along with the history of actions taken by the model. 

We now define the three actions\footnote{Except for the NONE argument where only two actions are applied: IGNORE and CONSTRUCT.} that the model applies at each time step to each argument, namely: add the argument (ADD), ignore the argument (IGNORE) and add the argument and construct an event candidate (CONSTRUCT). We have chosen only three actions for simplicity. Figure \ref{fig:search} shows how actions are applied to the argument \textit{angiogenesis} (due to space constraints we only show two actions indicated by the arrows). Concretely, we show two distinct structures ($S_1, S_2$) that are created in time step $t+1$ after \textit{CONSTRUCT} and \textit{IGNORE} actions were applied to the current structure $S_1$ in time step $i$.

The event candidate structures are fixed as events if the scores of the CONSTRUCT actions are above a certain threshold. The resulting state after a CONSTRUCT action is removed from the beams. We maintain multiple beams and use all of them to predict overlapping events.

\subsection{Neural Network as Scoring Function}%Classifier}
\label{nnmodel}

Figure \ref{fig:nn} shows the proposed neural model. We employ a \textbf{BiLSTM} network to generate the word representations from pre-trained \textbf{word embeddings}. To represent phrases, we averaged the word representations. The LSTM network is shared among the states during search in the sentence. We build a relation embedding for each argument, which concatenates the information of the trigger $t$, the role $o$, the argument $a$ and the action $c$. We include both the type information $p$ and the word or phrase representation $w$ of the trigger or entity argument. Formally, each relation $r_i$ is represented as a \textbf{relation embedding}: $\mathbf{r}_i =[\mathbf{t}_p; \mathbf{t}_w; \mathbf{o}_p; \mathbf{a}_p; \mathbf{a}_w;\mathbf{c}]$, where $\mathbf{t}_p$ is the representation of the type of the trigger and so on. 
Each $\mathbf{r}_i$ is passed to a linear \textbf{hidden layer}, then to a rectified linear unit (ReLU) non-linearity and summed to produce the \textbf{structure} and \textbf{buffer} embeddings: $\mathbf{S}_t$ and $\mathbf{B}_t$. 

We use a neural network as the \textbf{action scoring function} (indicated by the dotted box in Figure \ref{fig:nn}) defined as $\sigma({a}_t | {S}_{t-1}, {B}_{t-1})$. We model the action scoring function using $S_t$ and $B_t$, which are composed by adding an action ${a}_t$ for a relation $r_t$ to $S_{t-1}$ and moving $r_t$ from $B_{t-1}$ to $S_{t-1}$. The state at any time step $t$ is composed of the buffer $B_t$ and the partially built structure $S_t$. Each of $S_t$ and $B_t$ contains a set of relations ${\{r_1, r_2, r_3, \ldots, r_n\}}$. In Figure \ref{fig:nn}, there is no arrow to \textit{B} in event \textit{E1} because the diagram only shows the model snapshot at a specific time step during the search process. In this particular time step \textit{t}, the buffer \textit{B} in \textit{E1} is already empty and thus, it does not contain any relations $\mathbf{r}_i$.

$\mathbf{S}_t$ and $\mathbf{B}_t$ are then concatenated to form the \textbf{event embedding}. The event embedding has the same dimension as the sum of argument type and word dimensions so that it can be used as argument representation in nested events as shown in Figure \ref{fig:nn}. Then, we passed the event embedding into a linear \textbf{hidden layer} and output $z_t$. Finally, the scoring function $\sigma$ is calculated as $\sigma({a}_t | {S}_{t-1}, {B}_{t-1}) = \mathrm{sigmoid}({z}_t)$.

\subsection{Training}
\label{training}
% Since we will compare our model with the state-of-the-art  TEES event detection module~\cite{bjorne2018biomedical}, we train our model with the same input as theirs. Concretely, we merge the predicted relations from TEES with the pairwise relations decomposed from the gold events. We then generate a relation graph from the merged relations and 
From the relation graph generated in \S \ref{relgraph}, we calculate gold action sequences that construct the gold event structures on the graph. The loss is summed over all actions and for all the events during the beam search and thus the objective function is to minimise their negative log-likelihood. We employ early updates~\cite{collins2004incremental}: if the gold falls out of the beam, we stop searching and update the model immediately.

\section{Experimental Settings}

We applied our model to the BioNLP CG shared task 2013~\cite{pyysalo2015overview}. We used the original data partition and employed the official evaluation metrics. We focussed on the CG task dataset over other BioNLP datasets because of its complexity and size \cite{nedellec2013overview,bjorne2018biomedical,pyysalo2013overview}. The CG dataset has the most number of entity types and event types and thus is the most complex among the available (and accessible) BioNLP datasets. Furthermore, the CG dataset is the largest dataset in terms of the number of event instances and the proportion of nested and overlapping events. Evaluating our model extensively on other BioNLP tasks and datasets is part of our future work.
%, after removing illegal structures using the rules in the official evaluation script.

The development set contains 3,217 events of which 36.46\% are nested events, 43.05\% are overlapping events and 44.07\% are flat events. Note that the total does not equal to 100\% because nested and overlapping events may have intersection: a nested event can be overlapping and vice versa.
% We first separated the nested, overlapping and flat events, respectively. Then we compute the precision and recall for each category in the following way. For example, for nested events, to compute the precision, we compare the predicted nested events with all gold events and to compute recall, we compare gold nested events with all predicted events. The evaluation script detects nested events by comparing the whole tree structure down to its sub-events until it reaches the flat events. Thus, the performance scores of the nested events inevitably include the performance on flat events. 

We compared our model with the event detection module of the state-of-the-art model TEES~\cite{bjorne2018biomedical}, which employs convolutional neural network and uses syntactic and hand-engineered features for event detection. \citet{bjorne2018biomedical} found that the dependency parse features increased the performance of the convolutional model. In contrast, we do not use these syntactic features nor hand-engineered features. Furthermore, instead of the ensemble methods, we used TEES's published \textit{single} models, as this enables us to make a direct comparison with TEES in a minimal setting. %We cannot test our model on the test set gold annotation directly since it is withheld.
We train our model using the predicted relations from TEES merged with the pairwise relations decomposed from the gold CG events. During inference, we predict event structures using only the predicted relations from TEES.

\subsection{Nested and Overlapping Event Evaluation Process}
\label{nestedoverlapdetection}
Similarly, we used the official evaluation script to measure the performance of the model on nested, overlapping and flat events. We first separated the nested, overlapping and flat events, respectively. Then we compute the precision and recall for each category in the following way. For example, for nested events, to compute the precision, we compare the predicted nested events with all gold events and to compute recall, we compare gold nested events with all predicted events. The evaluation script detects nested events by comparing the whole tree structure down to its sub-events until it reaches the flat events. Hence, the performance scores of the nested events inevitably include the performance on flat events. 

\subsection{Training Details and Model Parameters}
\label{training_and_model_params}
We implemented our model using the Chainer library~\cite{chainer_learningsys2015}.  We initialised the word embeddings using pre-trained embeddings~\cite{chiu-EtAl:2016:BioNLP16} while other embeddings are initialised using the normal distribution. All the embeddings and weight parameters were updated with mini-batch using the AMSGrad optimiser~\cite{j.2018on}. We also incorporated early stopping to choose the number of training epochs and tuned hyper-parameters (dropout, learning rate and weight decay rate) using grid search. 
The model parameters can be found in \cref{params}. 

\section{Results and Analyses}\label{sec:results}

Table~\ref{perf} shows the event detection performance of the models on the test set. Our model achieves performance comparable to the state-of-the-art TEES event detection module without the use of any syntactic and hand-engineered features, suggesting it can be applied to other domains with no need for feature engineering. We validated it to have no significant statistical difference with the TEES model (the Approximate Randomisation test~\cite{yeh2000more,noreen1989computer}). 

To gain a deeper insight about the model, we performed analyses on the development set. Table~\ref{kbest} shows the performance of SBNN by varying the $k$-best parameter in beam search. We tested $2^i$ values for $i = {1,2,3,\ldots,11}$ and found that the best value was 8 with F1-score of 54.36\%, which is 2.2 percentage points (pp) higher than TEES.

\begin{table}[t!]
	\tiny
	\centering
	\resizebox{0.7\linewidth}{!}{%
	
	\begin{tabular}{llccc}
		\toprule[0.5pt]
% 		\multirow{2}[3]{*}{Model}  &  \multicolumn{3}{c}{Test Set (\%)}  \\
% 		\cmidrule(lr){2-4} 
		 Model & P & R & F1 (\%) \\
		\midrule[0.5pt]
        TEES            &61.42  & \textbf{52.93}  & 56.86 \\
		SBNN         & \textbf{63.67} &51.43 &\textbf{56.90}\\
		\bottomrule[0.5pt]
	\end{tabular}
    }
	\caption{Event detection performance on the CG task 2013 test dataset.}
	\label{perf}
\end{table}

\begin{table}[t!]
	\tiny
	\centering
	\resizebox{0.7\linewidth}{!}{%
	
	\begin{tabular}{llcccccc}
		\toprule[0.5pt]
		
		Model  & P & R & F (\%) \\
		\midrule[0.5pt]
		   TEES &56.81  &48.21  &52.16 \\
		\midrule[0.5pt]
        SBNN $k$ = 1   &48.95&	43.79&	46.23\\
		~~~~~~~~~~~~$k$ = 8 & \textbf{63.60}&	47.46 &	\textbf{54.36} \\
		~~~~~~~~~~~~$k$ = 64 & 60.30 &	\textbf{49.16} &	54.17 \\
		~~~~~~~~~~~~$k$ = 256 & 60.91	&48.53	&54.02\\

		\bottomrule[0.5pt]
	\end{tabular}
    }
	\caption{Event detection performance on CG task 2013 development set for SBNN with the top three performing $k$-values and when $k=1$.
	}
	\label{kbest}
\end{table}

Table~\ref{numclassifications} shows the number of classifications (or action scoring function calls in our model) performed by each model with the corresponding actual running time. SBNN performs fewer classifications and in less time than TEES, implying it is more computationally efficient.

Table~\ref{analysis} shows the performance comparison of the models on nested, overlapping and flat event detection. Our model yields higher F1-scores than TEES which can be attributed to its ability to maintain multiple beams and to detect events from all these beams during search. 

Finally, we computed the upper bound recall given the predicted relations from TEES. The upper bound is computed by setting the threshold parameter of our model to zero, which then constructs all gold events possible from the predicted relations of TEES. Since we evaluate our model on the output relations of TEES, the event detection performance is bounded or limited by these predicted relations. For instance, if one of the relations in an event was not predicted, the event structure will never be formed. We observe that this remains a challenging task since the upper bound recall is still at 53.47\% (6.01pp higher than our current model's score). Closing this gap requires among others addressing inter-sentence and self-referential events, which account for 3.1\% of the total events. 

\section{Conclusions and Future Work}
We presented a novel search-based neural model for nested and overlapping event detection by treating the task as structured prediction for DAGs. Our model achieves performance comparable to the state-of-the-art TEES event detection model without the use of any syntactic and hand-engineered features, suggesting the domain-independence of the model. Further analyses on the development set revealed some desirable characteristics of the model such as its computational efficiency while yielding higher F1-score performance. These results set the first focussed benchmark of our model and next steps include applying it to other event datasets in the biomedical and general domain. In addition, it can also be applied to other DAG structures such as nested/discontiguous
entities~\cite{
muis-lu:2016:EMNLP2016,
ju-miwa-ananiadou:2018:N18-1}. 

\begin{table}[t!]
	\tiny
	\centering
	\resizebox{0.9\linewidth}{!}{%
	
	\begin{tabular}{lrr}
		\toprule[0.5pt]
        Model & Number of Classifications & Running Time (s)\\
		\midrule[0.5pt]
		TEES        & 6,141 & 155\\
% 		Baseline    & 25,766\\
		SBNN $k$ = 8        & \textbf{4,093} & \textbf{131}\\
		\bottomrule[0.5pt]
	\end{tabular}
    }
	\caption{Comparison on computational efficiency on the CG task 2013 development dataset.}
	\label{numclassifications}
\end{table}
\begin{table}[t!]
	\tiny
	\centering
	\resizebox{0.9\linewidth}{!}{%
	
	\begin{tabular}{lcccc}
		\toprule[0.5pt]
        Model & Nested & Overlapping & Flat & Overall F1 (\%) \\
		\midrule[0.5pt]
		TEES        & 42.70 & 34.49 & 56.81 & 52.16\\
		SBNN  $k$ = 8         & \textbf{45.24} & \textbf{36.92} & \textbf{60.5} & \textbf{54.36}\\
		\bottomrule[0.5pt]
	\end{tabular}
    }
	\caption{Nested and overlapping event detection F1 (\%) score performance on the CG task 2013 development set.}
	\label{analysis}
\end{table}

\section*{Acknowledgement}
	We thank our anonymous reviewers for their invaluable feedback. This research was supported by funding from BBSRC Japan Partnering Award, Text mining and bioinformatics platforms for metabolic pathway modelling [Grant ID: BB/P025684/1] and AIRC/AIST. The first author gratefully acknowledges financial support from the University of the Philippines System Doctoral Studies Fund.

\bibliography{emnlp-ijcnlp-2019}
\bibliographystyle{acl_natbib}

\clearpage
\appendix

% \section{Baseline Model}
% \label{baseline}
% The baseline model enumerates exhaustively all the candidate structures from a given set of relations. We employed template matching, built from training data, as preprocessing to reduce the number of candidate structures.\footnote{The number is huge even after filtering as shown in
% \cref{sec:results}.} Finally, the structures are classified using the neural network as event or not. Since this model does not consider actions and search order including the order of arguments, we employ a Child-Sum Tree-LSTM to represent the candidate structures. We use relation representation $\mathbf{r}_i$ without actions as its input. 

\section{Model Parameters}
\label{params}
% We implemented our model using the Chainer library.  We initialised the word embeddings using pre-trained embeddings while other embeddings are initialised using the normal distribution. All the embeddings and weight parameters were updated with mini-batch using the AMSGrad optimiser. We also incorporated early stopping to choose the number of training epochs and tuned hyper-parameters (dropout, learning rate and weight decay rate) using grid search. Table \ref{tab:1} shows the hyper-parameters used in our experiments.
\begin{table}[th!]
  \centering
  \resizebox{\linewidth}{!}{%
  \begin{tabular}{lr}
  \toprule[0.5pt]
  Parameter & Value/Dimension \\
  \midrule[0.5pt]
%   \textit{Common Parameters} &\\
%   \midrule[0.5pt]
  Mini-Batch Size &100 \\
  Word Embedding & 200 \\
  BiLSTM Word Embedding &100 \\
  Role Type Embedding & 10 \\
  Trigger/Argument Type Embedding & 20\\
  Early Stopping Patience & 5 \\
  Dropout & 0.5 \\
  Learning Rate & 0.001\\
    Hidden Layer Size & 60 \\
    Event Embedding &100  \\
%   \midrule[0.5pt]
%   \textit{Baseline Model}&\\
%   \midrule[0.5pt]
%   I/O Embedding & 2\\
%   Event Prediction Threshold &0.5  \\
%   Weight Decay Rate & 0 \\
%   \midrule[0.5pt]
%   \textit{Search-based Model}&\\
%   \midrule[0.5pt]
    Action Score Threshold &0.5  \\
    Beam Size & 8 \\
  Action Embedding & 4\\
  Weight Decay Rate & 0.001 \\
    \bottomrule[0.5pt]
  \end{tabular}
  }
 \caption{Hyper-parameters used in our experiments.}
 \label{tab:1}
\end{table}

\end{document}